\documentclass[letterpaper, 10 pt, conference]{ieeeconf}  

\IEEEoverridecommandlockouts                              

\usepackage{graphicx}
\usepackage{booktabs}
\usepackage{multirow}
\usepackage{amsmath} 
\usepackage{subcaption}  
\usepackage{float}       
\usepackage{nicefrac}       

\title{\LARGE \bf
Learning-Based Distance Estimation for 360° Single-Sensor Setups
}

\author{Yitong Quan$^{1}$, Benjamin Kiefer$^{12}$, Martin Messmer$^{1}$ and Andreas Zell$^{1}$
\thanks{$^{1}$University of Tuebingen, 
        {\tt\small first.last@uni-tuebingen.de} \hphantom{iii} $^{2}$LOOKOUT, {\tt\small benjamin@lookout.team}}%
}

\begin{document}

\maketitle
\thispagestyle{empty}
\pagestyle{empty}

\begin{abstract}

Accurate distance estimation is a fundamental challenge in robotic perception, particularly in omnidirectional imaging, where traditional geometric methods struggle with lens distortions and environmental variability. In this work, we propose a neural network-based approach for monocular distance estimation using a single 360° fisheye lens camera. Unlike classical trigonometric techniques that rely on precise lens calibration, our method directly learns and infers the distance of objects from raw omnidirectional inputs, offering greater robustness and adaptability across diverse conditions. We evaluate our approach on three 360° datasets (LOAF, ULM360, and a newly captured dataset Boat360), each representing distinct environmental and sensor setups. Our experimental results demonstrate that the proposed learning-based model outperforms traditional geometry-based methods  and other learning baselines  in both accuracy and robustness. These findings highlight the potential of deep learning for real-time omnidirectional distance estimation, making our approach particularly well-suited for low-cost applications in robotics, autonomous navigation, and surveillance.

\end{abstract}

\section{INTRODUCTION}

Accurate distance estimation is a fundamental requirement for a wide variety of perception tasks, including autonomous vehicle navigation, robotic manipulation, and surveillance. In particular, accurately determining how far away an object lies becomes essential for safe decision-making and interaction in dynamic environments. Although stereo or multi-camera setups and LiDAR systems have been adopted in many industrial applications, their high cost and complexity can be prohibitive in resource-constrained scenarios. Monocular depth estimation with a single pinhole camera is more affordable but suffers from a limited field of view (FOV). By contrast, using a single 360° fisheye camera offers a more affordable, lightweight, and comprehensive way to capture the surrounding scene \cite{yang2023large,quan2024robust}.

However, the extreme distortion inherent in fisheye lenses poses unique challenges for distance estimation. Classical methods typically rely on carefully calibrated lens models and trigonometric formulas to project 2D image coordinates into 3D space \cite{quan2024robust}. These geometry-based approaches perform well under stable camera intrinsics and extrinsics, lighting, minimal occlusion, and uniform lens characteristics. Yet, any deviation from assumed calibration conditions, such as change in camera extrinsics or intrinsics, unmodeled distortions or dynamic lighting, can yield substantial errors in practice \cite{genovese2024single}. Moreover, adopting these methods across different camera models, scene types, and environmental conditions often requires extensive recalibration.

Conventional trigonometric and polynomial-based lens models offer a mathematically elegant solution for fisheye image projection. Nevertheless, their accuracy relies on the completeness of the lens distortion parameters and the static nature of the environment. For instance, once the height of the camera or the angle of the lens changes, the precomputed calibration parameters may no longer be valid. Similarly, the presence of reflective surfaces, moving platforms, or occlusions can significantly degrade the performance of purely geometric methods. Furthermore, these approaches often treat each image frame independently, lacking the capacity to learn and refine distance estimations from large, diverse datasets.


\begin{figure}[t]
\centering
\begin{minipage}[t]{0.246\textwidth}
  \centering
  \includegraphics[trim=0 0 0 0,clip,width=\linewidth]{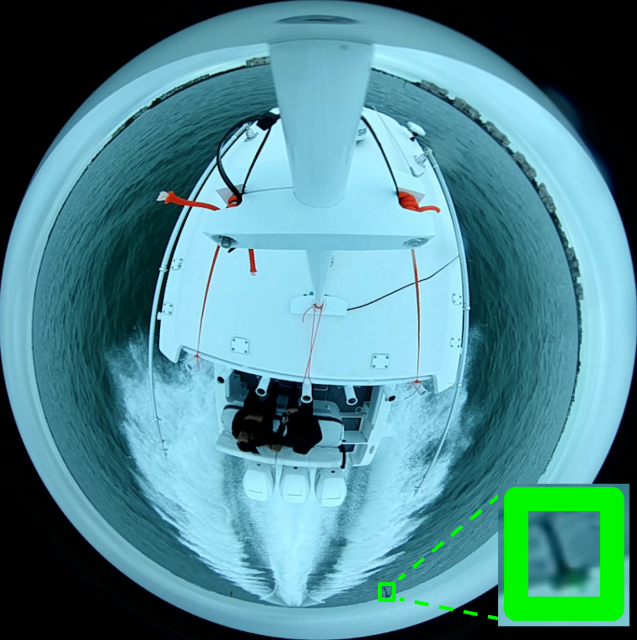}
\end{minipage}\hfill
\begin{minipage}[t]{0.238\textwidth}
  \centering
  \includegraphics[trim=0 0 0 0,clip,width=\linewidth]{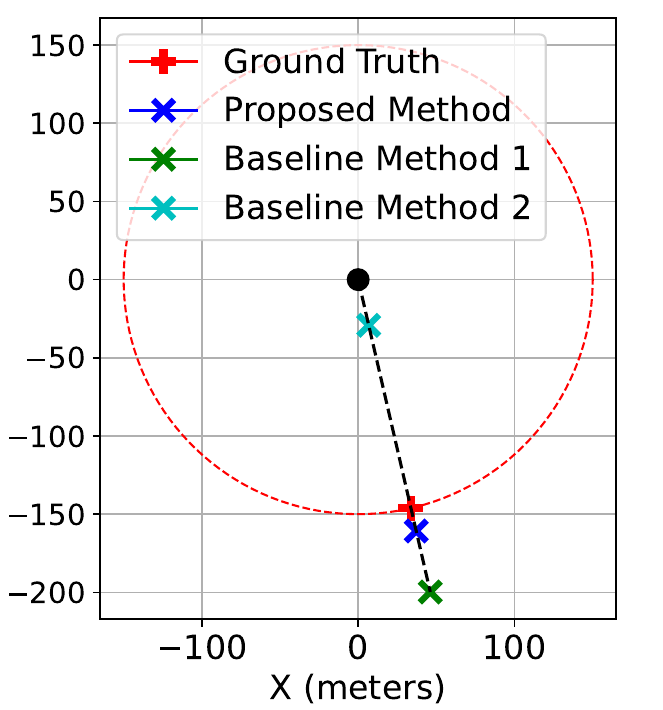}
\end{minipage}
\caption{As the boat accelerates, our learning-based method estimates 165m, closer to the ground truth (150m) than the geometry method (205m) and DepthAnythingV2 \cite{yang2025depth} (30m), showing robustness to camera pose changes in dynamic environments.}
\vspace{-0.5cm}
\label{fig:boat_acc}
\end{figure}

Advances in deep learning have catalyzed a shift toward data-driven solutions for visual perception tasks, including monocular depth estimation in regular perspective images. Without explicit geometric formulas, neural networks learn mappings between pixel intensities and depth values from training data alone. Although research on omnidirectional depth estimation is less extensive than on standard pinhole cameras, the potential for a learning-based system to adapt to fisheye distortions and dynamic environments is immense. Using convolutional architectures, these methods can capture high-level context and local texture information to predict distances more robustly than purely trigonometric models.

In summary, our key contributions are as follows:
\begin{itemize}
    \item \textbf{A lightweight yet highly effective learning-based network} for real-time object distance estimation in omnidirectional images.
    \item \textbf{Extensive empirical validation}, demonstrating the superiority of our approach over both geometric and deep learning baselines through testing on multiple datasets, including real-world deployment scenarios.
    \item \textbf{Benchmark dataset contribution}, providing  high-quality, annotated datasets for omnidirectional distance estimation to facilitate future research upon paper acceptance.

\end{itemize}


\section{RELATED WORK}


Distance estimation is a fundamental aspect of numerous perception tasks, including robotic manipulation, autonomous vehicle navigation, and surveillance. 
Traditional stereo vision techniques, such as Semi-Global Matching \cite{hirschmuller2007stereo} and deep learning-based methods like PSMNet \cite{chang2018pyramid}, provide dense depth estimation but require well-calibrated stereo pairs and suffer in textureless or occluded regions. On the other hand, LiDAR-based methods \cite{chen2017multi} offer highly accurate depth measurements and are widely used in autonomous driving, but they are expensive and resource intensive. 

Monocular 360° fisheye cameras present an affordable and comprehensive solution for capturing full environmental context, but the distortions introduced by such optics pose unique challenges for depth estimation.
Traditional geometry-based techniques rely on calibrated lens models and trigonometric formulas to project 2D image coordinates into 3D space \cite{yang2023large, quan2024robust}. 
However, the reliance on precise lens calibration makes them less adaptable to diverse environmental conditions and real-world distortions, as highlighted by \cite{son2023monocular}.

Advances in deep learning have enabled data-driven approaches to monocular depth estimation, which learn mappings between pixel intensities and depth values directly from training data. Eigen et al. introduced a multi-scale deep network for depth map prediction, demonstrating the potential of learning-based methods to extract meaningful depth information from single images \cite{eigen2014depth}. These techniques have since evolved to incorporate self-supervised learning, as seen in Godard et al.'s work, which introduces a left-right consistency loss to improve monocular depth estimation \cite{godard2017unsupervised}.

Adapting learning-based methods to omnidirectional images presents additional challenges due to fisheye distortions and non-uniform image projections. Rey-Area et al. proposed 360MonoDepth  \cite{rey2022360monodepth}, which projects 360° images onto tangent planes to create perspective views that can be processed using existing depth estimation models. The individual depth predictions are then merged using deformable multi-scale alignment and gradient-domain blending to construct a dense, high-resolution 360° depth map. Similarly, Yan et al. introduced FisheyeDistill  \cite{yan2022fisheyedistill}, which employs an ordinal distillation loss to transfer depth information from a teacher model, enhancing performance on fisheye images. Li et al. developed OmniFusion \cite{li2022omnifusion}, a method that transforms omnidirectional images into less-distorted perspective patches and integrates 3D geometric priors with 2D image features using a transformer-based architecture. Kumar et al. proposed UnRectDepthNet \cite{kumar2020unrectdepthnet}, a self-supervised framework that eliminates the need for image rectification, preserving the full field of view while reducing computational costs during inference. 
Zhang et al. proposed Depth-Anything-V2 \cite{yang2025depth}, a large-scale general-purpose depth estimation model designed to handle diverse scene variations. Yet, its reliance on extensive supervised training makes it unsuitable for scenarios where pixel-wise depth annotations are unavailable. Furthermore, its high computational cost limits its feasibility for real-time  deployment in autonomous navigation.

Applications of these methods in robotics and autonomous systems have demonstrated substantial benefits, particularly in object geolocalization and obstacle detection. Kumar et al. introduced SynDistNet \cite{kumar2021syndistnet}, which integrates monocular depth estimation with semantic segmentation, enhancing depth edges and object localization accuracy. This synergy is critical for autonomous navigation and robotic perception, where accurate distance estimation directly influences decision-making and real-time obstacle avoidance.

Our approach differs from existing methods by emphasizing object-based depth estimation, enabling real-time performance suitable for deployment in resource-constrained environments. By optimizing inference efficiency, we achieve accurate depth predictions while maintaining computational feasibility, making our system particularly well-suited for robotics and surveillance applications.

While traditional geometry-based methods retain their relevance for controlled environments and well-calibrated setups, learning-based approaches offer superior adaptability and robustness, particularly for real-world scenarios where lens distortions and environmental variations cannot be easily corrected. The discussed neural network-based methods highlight the potential of deep learning in advancing monocular distance estimation for omnidirectional imagery, making them strong candidates for deployment in robotic perception, surveillance, and autonomous navigation applications.

\section{METHODOLOGY}

\begin{table*}[ht]
\centering
\caption{Dataset Characteristics}
\label{tab:dataset_comparison}
\begin{tabular*}{\textwidth}{@{\extracolsep{\fill}} lcccc c}
\toprule
Dataset & Platform & Camera Mounting & Distance GT & Environment & Image Resolution \\ 
\midrule
LOAF & Static & Fixed height (varied per scene) & Manual with reference ruler & Indoor \& Outdoor & 512 × 512 \\
ULM360 & Moving vehicle & Mounted on a pole & LiDAR-based annotation & Urban & 1920 × 1920 \\
BOAT360 & Moving boat & Mounted on a mast & Chart Data & Open water & 1452 × 1452 \\
\bottomrule
\end{tabular*}
\end{table*}

\subsection{Datasets}
\label{subsec:Datasets}
We use three datasets, all featuring large-FOV ($>180^\circ$) downward-facing fisheye cameras but collected under different conditions to test the generalization and robustness of our approach, as summarized in  Table \ref{tab:dataset_comparison}.

\begin{figure*}[htbp]
    \centering
    \begin{subfigure}{0.195\textwidth}
        \includegraphics[width=\linewidth]{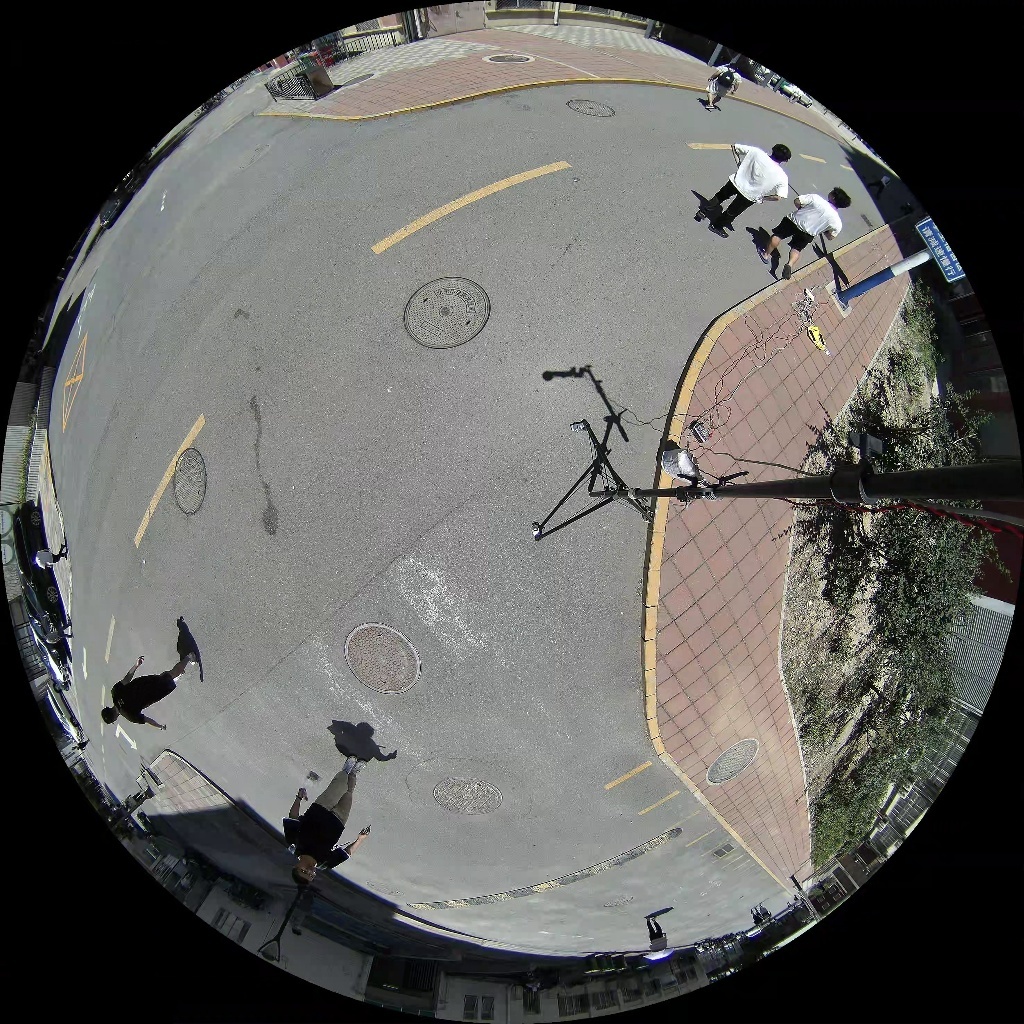}
        \caption{}
        \label{fig:sub1}
    \end{subfigure}
    \begin{subfigure}{0.78\textwidth}
        \centering
        \includegraphics[width=\linewidth]{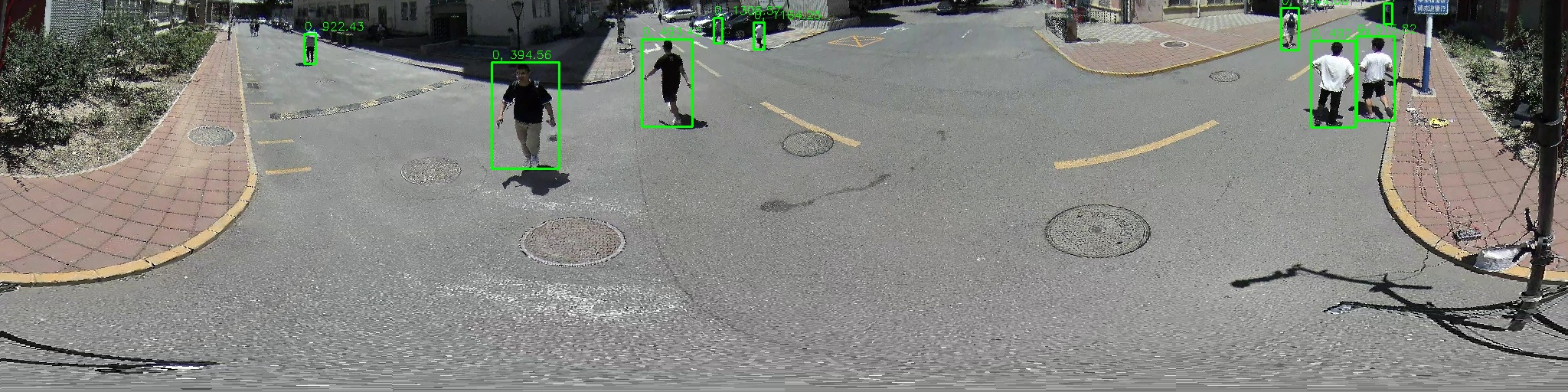}
        \caption{}
        \label{fig:sub6}
    \end{subfigure}

    \begin{subfigure}{0.195\textwidth}
        \includegraphics[width=\linewidth]{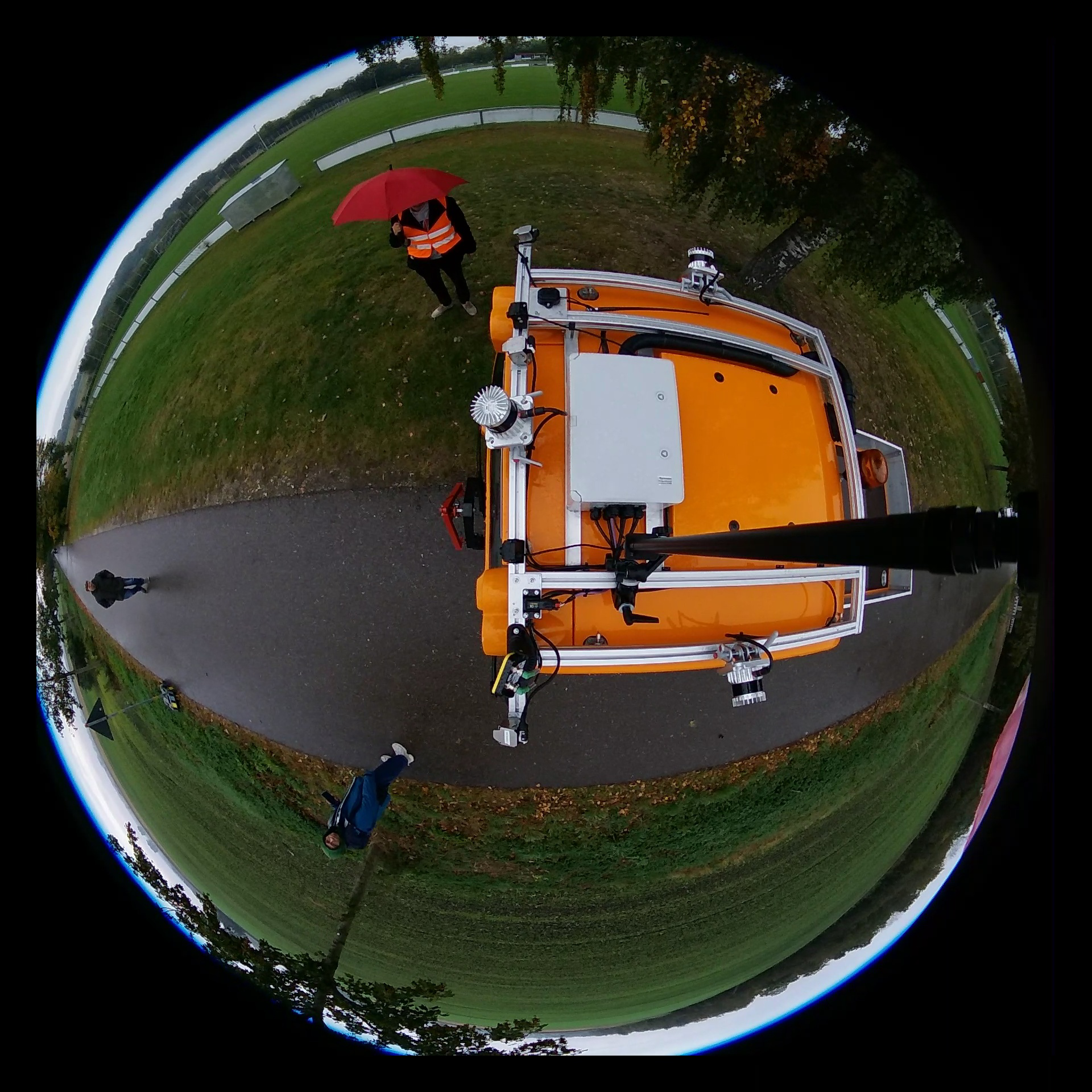}
        \caption{}
        \label{fig:sub1}
    \end{subfigure}
    \begin{subfigure}{0.78\textwidth}
        \centering
        \includegraphics[width=\linewidth]{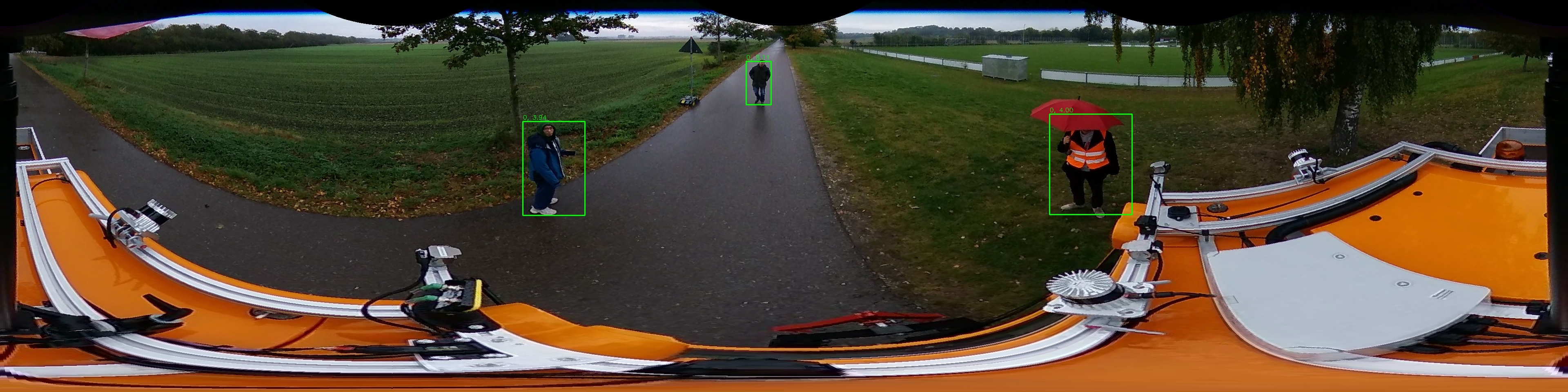}
        \caption{}
        \label{fig:sub6}
    \end{subfigure}

    \begin{subfigure}{0.195\textwidth}
        \includegraphics[width=\linewidth]{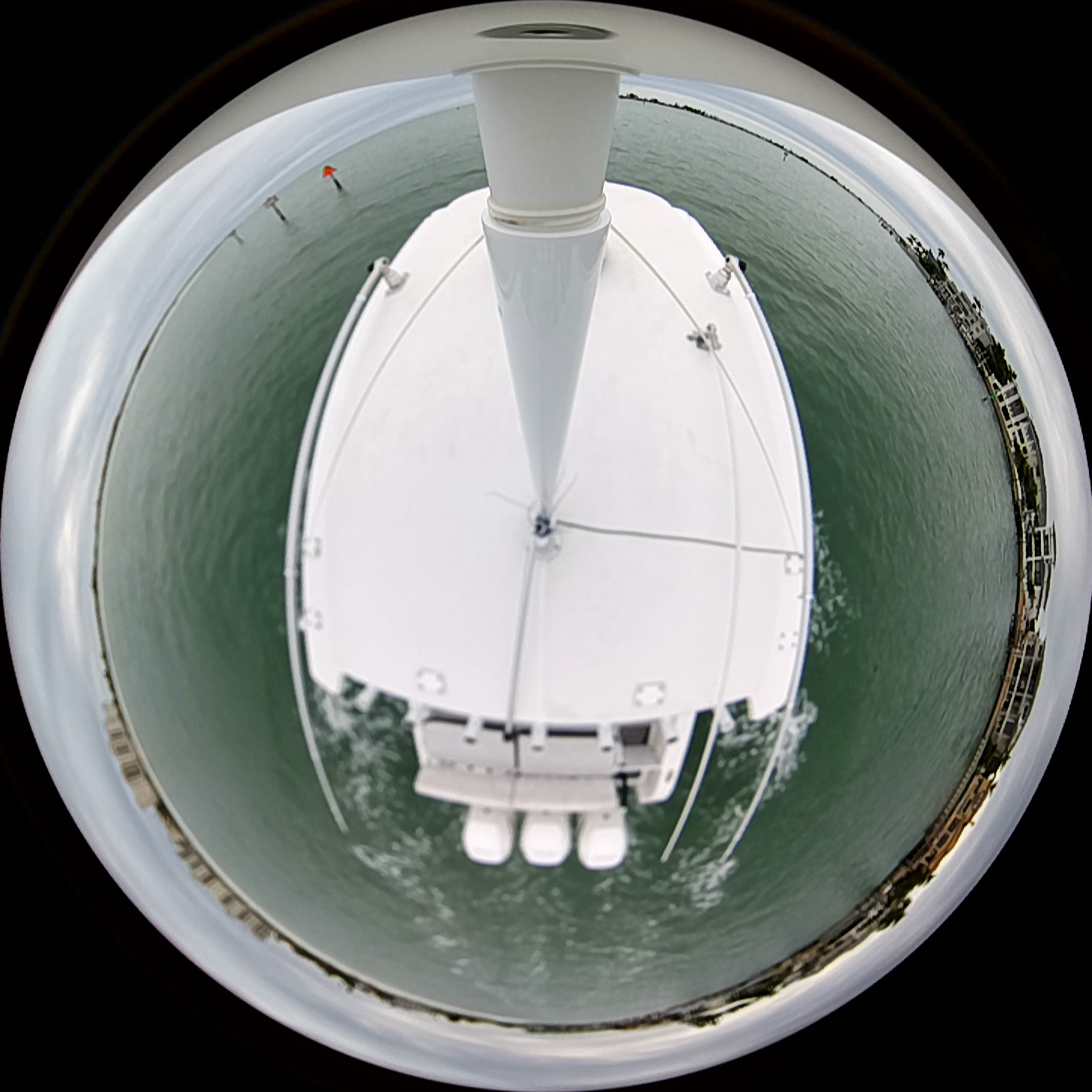}
        \caption{}
        \label{fig:sub1}
    \end{subfigure}
    \begin{subfigure}{0.78\textwidth}
        \centering
        \includegraphics[width=\linewidth]{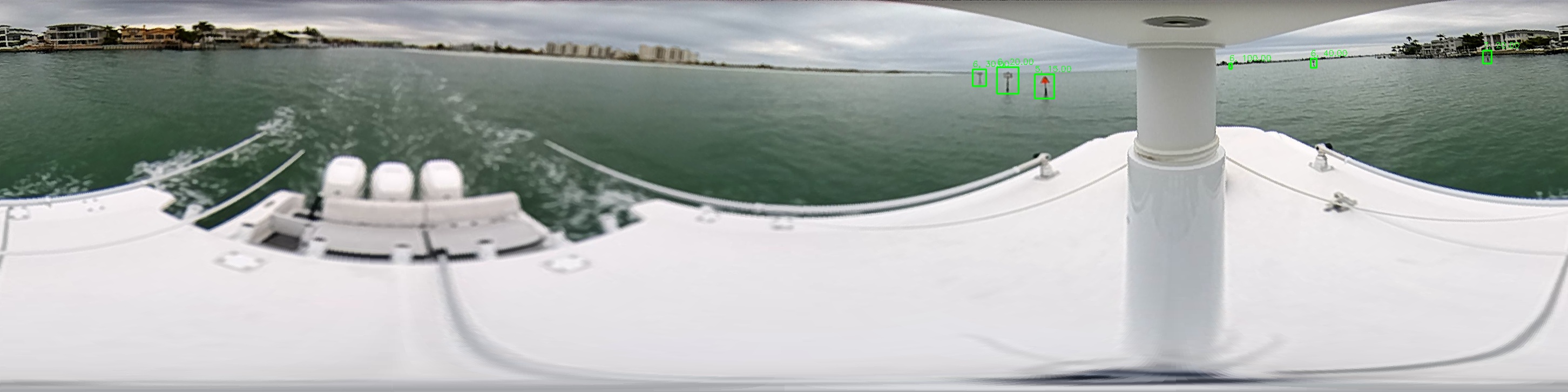}
        \caption{}
        \label{fig:sub6}
    \end{subfigure}
  
    \caption{(a, c, e) shows samples of raw images  from dataset LOAF, ULM360, BOAT360. Their equirectangular-transformed images and the corresponding annotations are shown as (b, d, f).}
    \label{fig:multirow_figure}
\end{figure*}

\begin{itemize}
    \item \textbf{LOAF (public dataset)}: Captured from a static camera setups at various heights. Distance ground truth is established using a  reference ruler, with one end positioned directly beneath the fisheye camera to serve as a calibration reference. Each annotated person in the dataset includes a bounding box along with the manually measured distance information \cite{yang2023large}.
    \item \textbf{ULM360 (our dataset)}: Collected in Ulm, Germany, using a mobile vehicle equipped with a downward-facing fisheye camera mounted on a pole. To provide precise ground-truth distances, the vehicle was equipped with three Ouster LiDARs, which allowed us to map each manually labeled bounding box to its corresponding point cloud distance \cite{quan2024robust}.
    \item \textbf{BOAT360 (our dataset)}: Collected from a moving boat, with the fisheye camera mounted on a mast. Unlike the previous datasets, where distances were measured using structured references (rulers or LiDARs), object distances in BOAT360 were estimated by experienced sailors. This dataset presents additional challenges due to boat dynamics, lack of static reference points, and dynamic environmental conditions.
\end{itemize}

Each dataset has different size and resolution, providing an opportunity to evaluate the generalization ability of our method across diverse real-world conditions. Additionally, we generate equirectangular representations of both the original captured images and their corresponding annotated bounding boxes. This ensures compatibility with alternative image processing techniques and allows us to evaluate the generalization capabilities of our method.



\subsection{Model Architecture}

\begin{figure*}
\centering
\includegraphics[width=1\textwidth]{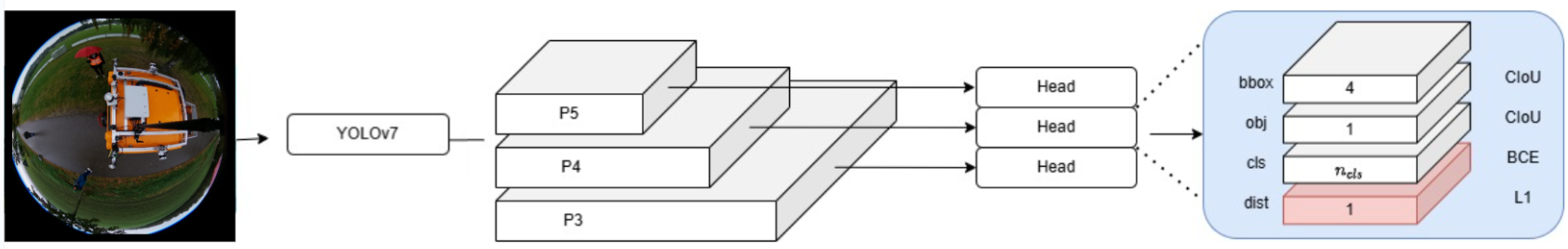}
\caption{Architecture of our proposed approach, illustrated using YOLO models. We integrate a dedicated distance estimation branch (red) with tailored feature transformations and loss formulations, enabling simultaneous object detection and accurate depth estimation while maintaining real-time performance.}
\label{fig:yolodistances}
\end{figure*}

For our task, we utilize the YOLOv7 object detection architectures, which are recognized for their very good balance between accuracy and real-time performance, positioning them on the Pareto front of efficiency \cite{wang2023yolov7}. To enable these models to estimate object distances, we extend their architecture by incorporating an additional output neuron per anchor, responsible for predicting the distance, as illustrated in Figure \ref{fig:yolodistances}.

\subsection{Distance Normalization Strategies}

Integrating this additional output into the detection framework allows the model to estimate object distances simultaneously with object detection. However, a direct regression of metric distances poses challenges due to the natural tendencies of these networks to predict values that are centered around zero and of relatively low magnitude \cite{wang2023yolov7}. Training instability and performance degradation can occur when predicting raw distances. To address this, we experimented with two normalization techniques for the distance estimation branch: linear scaling and logarithmic transformation tailored to different distance distributions and ranges.

\subsubsection{Linear Normalization}

In linear normalization, the actual distance \(d\) is scaled to a normalized value \(y\) within the range \([0,1]\).




\subsubsection{Logarithmic Normalization}

A logarithmic transformation is applied before normalization to better handle variations across a wide range of distances. This allows the model to penalize errors in the estimation of closer objects more heavily compared to objects farther away. The transformation is given by:

\begin{equation}
y = \frac{\log(d + \epsilon)}{\log(d_{\text{max}} + \epsilon)}
\end{equation}

where   \(d_{max}\) is a predefined maximum distance,  and \(\epsilon\) is a small constant  to prevent numerical instability when \(d = 0\). The original distance is then recovered during inference as:

\begin{equation}
\hat{d} = \exp\left( y \cdot \log(d_{\text{max}} + \epsilon) \right) - \epsilon
\end{equation}











These normalization strategies ensure that distance predictions remain well-scaled for neural network outputs. During training, the model minimizes the loss between the normalized predictions and ground truth distances. In inference, the predicted values are mapped back to real-world distances using the respective inverse transformations. Our experiments revealed no significant performance differences between these strategies. Due to its simplicity, we selected linear normalization for subsequent experiments.

\subsection{Model Training}

The model is trained using a multi-component loss function that balances different objectives:

\begin{itemize}
  \item Objectness Loss: Encourages correct detection of object presence.
  \item  Classification Loss: Ensures accurate identification of detected objects.
  \item Localization Loss: Measures how precisely bounding boxes match ground truth.
  \item Distance Loss: Optimizes the accuracy of predicted distances.
\end{itemize}

The first three loss functions remain unchanged from the original YOLO framework, while the distance loss is computed using the \(L_1\) norm.

The total loss function is formulated as:

\begin{equation}
\mathcal{L}_{total} = \lambda_{obj} \mathcal{L}_{obj} + \lambda_{cls} \mathcal{L}_{cls} + \lambda_{loc} \mathcal{L}_{loc} + \lambda_{dist} \mathcal{L}_{dist}
\label{eqn:loss_function}
\end{equation}

where $\lambda_{obj}$, $\lambda_{cls}$, $\lambda_{loc}$, and $\lambda_{dist}$ are weighting coefficients that control the contribution of each loss term. The distance loss is defined as:

\begin{equation}
\mathcal{L}_{dist} = \frac{1}{N} \sum_{i=1}^{N} \left| \hat{d}_i - d_i \right|
\end{equation}

where $N$ is the number of detected objects, $\hat{d}_i$ is the predicted distance for object $i$, and $d_i$ is the ground truth distance.

\section{EVALUATION METRICS}
\label{sec:evaluation_metrics}

To assess our model's performance in both object detection and distance estimation, we use mean Average Precision (mAP) \cite{cocodataset} at different Intersection over Union (IoU) thresholds \cite{recall-IoU} for  object detection, and assess distance estimation using both absolute and confidence-weighted error metrics, which defined as following.

\subsection{Absolute Distance Error}

The absolute distance error measures the mean absolute difference between predicted and ground truth distances, quantifying the discrepancy for detected objects. It directly evaluates distance prediction accuracy but does not consider the model’s confidence in each detection.






\subsection{Confidence-Weighted Distance Error}

To incorporate the model's confidence in its predictions, we introduce a confidence-weighted distance error metric. This metric ensures that errors from high-confidence detections contribute more significantly to the final evaluation, aligning the error measurement with the model’s certainty.

 Given a set of \( M \) detected objects, for each  object \( i \), let \( c_i \) denote the confidence score assigned by the model. The confidence-weighted distance error \( E_w \) is computed as:

\begin{equation}
E_w = \frac{\sum_{i=1}^{M} c_i \cdot |d_i - \hat{d}_i|}{\sum_{i=1}^{M} c_i} .
\end{equation}

Unlike standard absolute error metrics, this formulation down-weights the impact of low-confidence detections while emphasizing the accuracy of high-confidence predictions. As a result, it provides a more reliable assessment of distance estimation performance, particularly in real-world scenarios where model confidence is crucial for decision-making.

Following the standard practice in mAP calculations to ensure meaningful evaluation, distance errors are only considered for detections that meet a predefined IoU threshold with ground truth objects, filtering out false positives. This approach ensures both the absolute distance error and confidence-weighted distance error reflects errors for valid detections rather than spurious predictions.

\begin{table*}[h]
\centering
\caption{Object detection and distance estimation performance on LOAF dataset (baselines italicized).}
\scriptsize
\begin{tabular*}{\textwidth}{@{\extracolsep{\fill}} lcccccc}
\toprule
Model & Weighted Error & Absolute Error (m) & Precision & Recall & mAP@50 & mAP@50:95 \\
\midrule
\textit{Y7-{\text{w/oDist}}} & - & - & 0.617 & 0.496 & 0.409 & 0.149 \\
\textit{Y7-Geo} & 1.203 & 10.822 & 0.653 & 0.502 & 0.438 & 0.169 \\
\textit{Y7-DA2 }& 0.858 & 10.950 & 0.653 & 0.502 & 0.438 & 0.169 \\
Y7-D & 0.070 & \textbf{0.874} & 0.653 & 0.502 & 0.438 & 0.169 \\
Y7-D-Eq & 0.210 & 1.609 & \textbf{0.799} & \textbf{0.601} & \textbf{0.631} & \textbf{0.274} \\
Y7-D+C & \textbf{0.066} & \textbf{0.874} & 0.690 & 0.531 & 0.495 & 0.229 \\
\bottomrule
\end{tabular*}
\vspace{-0.3cm}
\label{tab:LOAF_overall}
\end{table*}

\section{EXPERIMENTS AND RESULTS}

This section presents an analysis of the experimental results, focusing on evaluating the proposed learning-based distance estimation approach. The key research questions addressed include:

\begin{enumerate}
    \item Whether adding a distance estimation head affects object detection accuracy.
    \item Whether our proposed learning-based model outperforms a well-calibrated geometric model and other learning-based approaches.
    \item Whether our proposed models are more robust to changes in camera parameters compared to geometric methods.
    \item Whether training directly on fisheye images is better than using equirectangular projections. The question arises because fisheye images preserve raw spatial distortions that might aid depth estimation, while equirectangular projections provide axis-aligned bounding boxes that could better benefit pretrained object detection models.
    \item Whether incorporating another camera height estimation head improves distance prediction. This question arises because camera height is a crucial parameter in geometric distance estimation, and learning it jointly with object detection may provide additional supervisory signals that improve depth prediction.
\end{enumerate}

\begin{figure}[ht]
\centering
		\includegraphics[trim=0 0 0 35,clip,width=.205\textwidth]{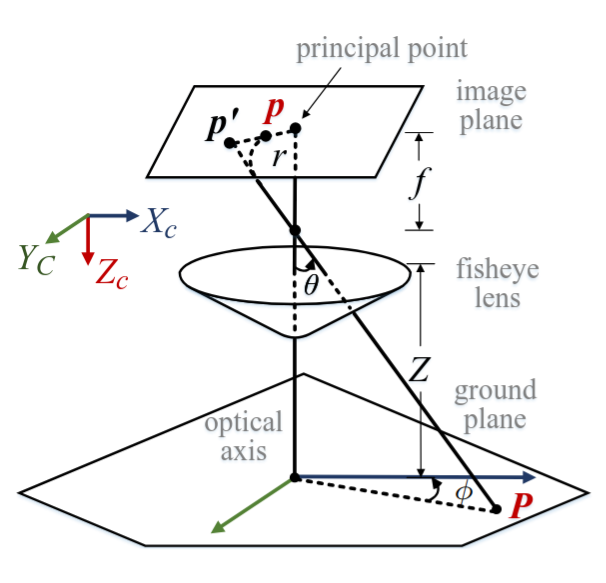}
  \includegraphics[trim=0 0 0 0,clip,width=0.27\textwidth]{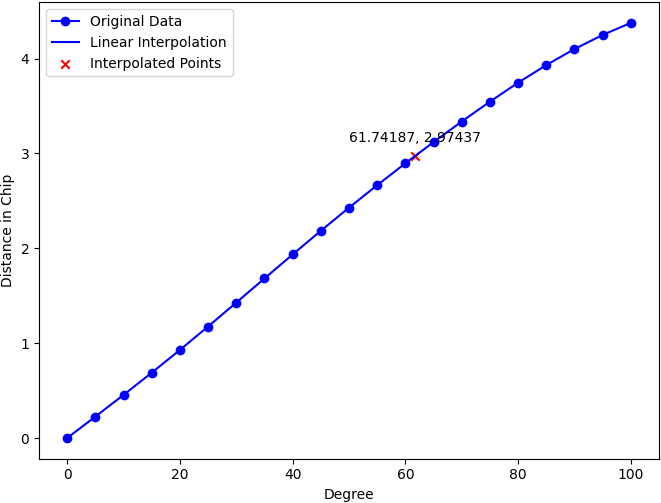}
\caption{(Left) The fisheye lens model for geometric distance calculation.  (Right) The interpolation between $\theta$  with respect to the optical axis and $r$ in millimeters on the chip. \cite{quan2024robust}}
	\label{fig:img_lens_model}
\end{figure}

To achieve this, we consider multiple model configurations, each designed to investigate different aspects of object detection and distance estimation. The following models are employed in our study:

\begin{itemize}
    \item \textit{Y7-{\text{w/oDist}}}: A baseline object detection model Yolov7 without the distance estimation head. In this configuration, the distance loss weight is set to zero, meaning the network is trained solely for object detection.

    \item \textit{Y7-{\text{Geo}}}: A baseline model that utilizes a geometric distance calculation approach based on a well-calibrated fisheye camera model. Given the pixel location of the detected bounding box predicted by Yolov7, the model estimates distance using precomputed geometric transformations, as illustrated in Figure~\ref{fig:img_lens_model}.

    \item \textit{Y7-DA2}: A baseline model using Depth-Anything-V2 (small) \cite{yang2025depth} estimates pixel-wise depth and averages the n lowest pixel depths within each detected bounding box to determine the object's distance.
    
    \item \textit{Y7-D}: A model that extends the baseline \textit{Y7-{\text{w/oDist}}} by incorporating a dedicated distance estimation head. This model jointly optimizes both object detection and distance estimation in an end-to-end manner.

    \item \textit{Y7-D-Eq}: Similar to \textit{Y7-D}, but trained on equirectangular projections of fisheye images rather than the original fisheye format. This approach aims to assess the impact of input representation on detection and distance estimation accuracy.
    
    \item \textit{Y7-D+C}: A model that extends \textit{Y7-D} by adding an auxiliary head for estimating the camera height. The inclusion of camera height estimation aims to provide additional geometric constraints, potentially improving the model’s depth estimation performance.
\end{itemize}

All models were trained on their respective datasets for 200 epochs. When training with fisheye images, augmentations such as rotation and flipping were applied, while transformations that could distort the fisheye structure, such as scaling, shearing, and perspective changes, were disabled. For training with equirectangular images, only flipping and color adjustments were used. In the case of \textit{Y7-D} and \textit{Y7-D-Eq}, the weight coefficient \( \lambda_{\text{dist}} \) in Equation (\ref{eqn:loss_function}) is empirically set to \( \nicefrac{1}{3} \), while the remaining \( \nicefrac{2}{3} \) is distributed among the other loss components preserving their default ratios from \cite{wang2023yolov7}. For \textit{Y7-D+C}, \( \lambda_{\text{dist}} \) is empirically set to \( \nicefrac{1}{6} \), with an additional $ \lambda_{\text{ad}} = \nicefrac{1}{6} $ assigned to the loss of height estimation of the camera.  The remaining hyperparameters for model training follow the default settings in \cite{wang2023yolov7}.

We first evaluate our models on the LOAF dataset and subsequently analyze its generalizability across the ULM360 and BOAT360 datasets. Additionally, we evaluate the real-time inference capability of our model on an NVIDIA Orin AGX using 1024×1024 resolution images, achieving a processing speed of 45 FPS, as shown in Table \ref{tab:inference_speed}, demonstrating its suitability for real-time applications.

\subsection{Evaluation on LOAF Dataset}

\subsubsection{Impact of Distance Estimation Head on Object Detection}

A crucial aspect of our approach is integrating a distance estimation head into an object detector. A key concern is whether this additional head negatively impacts detection performance. To analyze this, we compare \textit{Y7-{\text{w/oDist}}} against \textit{Y7-D}.


As seen in Table~\ref{tab:LOAF_overall}, adding the distance estimation head results in a slight decrease in precision but an increase in recall and an improvement in mAP@50. These results indicate that integrating a distance estimation head does not significantly degrade object detection accuracy, making it feasible to use a unified network for object detection and distance estimation.

\subsubsection{Comparison to Geometric Methods and other Learning-Based Models}

To evaluate the effectiveness of our learning-based approach, we compare the performance of \textit{Y7-D} against both \textit{Y7-Geo} and \textit{Y7-DAv2}.
From Table~\ref{tab:LOAF_overall}, the geometric model \textit{Y7-Geo} performs significantly worse, with an absolute error of 10.747 m compared to 0.874 m for our proposed model. Additionally, the weighted error for \textit{Y7-Geo} is 1.153, while \textit{Y7-D} achieves a significantly lower error of 0.070. These results show that our proposed  learning-based models generalize better than geometric models, which rely on precise camera calibration and extrinsic parameters.

Compared to the othe learning-based baseline \textit{Y7-DAv2}, \textit{Y7-D} also achieves higher accuracy and efficiency. 
Since our dataset lacks pixel-wise depth annotations, \textit{Y7-DAv2} could not be fine-tuned, likely leading to its suboptimal performance. However, the high computational complexity of \textit{Y7-DAv2} makes it impractical for real-time applications, as in Table \ref{tab:inference_speed}.

\begin{table}[t]
\centering
\scriptsize
\caption{Inference speed (FPS) on  1024×1024 resolution images.}
\label{tab:inference_speed}
\begin{tabular}{l cc}
\toprule
Model &   RTX3090&   NVIDIA Orin AGX \\
\midrule
Y7-D / Y7-Geo  / Y7-D-Eq  / Y7-D+C      & \textbf{166} & \textbf{45} \\
Y7-DA2      &  39 & 10\\
\bottomrule
\end{tabular}
\vspace{-0.5cm}
\end{table}

\subsubsection{Robustness to Changes of Camera Parameters}

According to \cite{handel2007analyzing}, the image could drift by up to several pixels during the camera warm-up process, which correlates with temperature-induced shifts in the principal point.  To assess the robustness of our approach under such perturbations, we simulate this scenario by shifting the image along the x-axis by one and five pixels and analyzing the resulting distance estimation errors. Table \ref{tab:pxl_offset_distance_errors} summarizes the impact of these shifts on both the geometric baseline and our proposed learning-based method.

\begin{table}[ht]
\centering
 \scriptsize
 \vspace{-0.0cm}
\caption{Robustness against principal point shifts. The table compares Absolute Error  for Y7-Geo (baseline) and Y7-D (proposed method) on LOAF dataset under principal point offset.}
\label{tab:pxl_offset_distance_errors}
\begin{tabular*}{\columnwidth}{@{\extracolsep{\fill}} ccccc}
\toprule
\multirow{2}{*}{Offset} & \multicolumn{2}{c}{Y7-Geo Abs.Err (m)} & \multicolumn{2}{c}{Y7-D Abs.Err (m)} \\ 
\cmidrule(lr){2-3} \cmidrule(lr){4-5}
&range 28.8 - 36.0 m  & Overall & range 28.8 - 36.0 m & Overall\\ 
\midrule
No offset & 14.590 & 10.822 & 3.490 & 0.874 \\
1 pixel offset & 14.977 & 10.891 & 3.649 & 0.896 \\
$\Delta$ & 0.387 & 0.069 & \textbf{0.159} & \textbf{0.022} \\ 
\midrule
5 pixel offset & 15.606 & 10.929 & 3.827 & 0.920 \\
$\Delta$ & 1.016 & 0.107 & \textbf{0.337} & \textbf{0.046} \\ 
\bottomrule
\end{tabular*}
\end{table}

With a one-pixel offset, the geometric method, \textit{Y7-Geo}, already exhibits a small but noticeable increase in overall absolute error (+0.069 m). In contrast, our learning-based approach, \textit{Y7-D}, shows only a 0.022 m rise in absolute error, indicating strong resilience to small perturbations. As the shift increases to five pixels, the performance gap between the two methods widens further, our model maintains lower error growth, confirming its robustness against  minor optical misalignments. The increase in absolute error is more severe when estimating far-distance objects in the range of 28 m to 36 m, where geometric methods deteriorate significantly.

\begin{table}[ht]
\centering
 \scriptsize
\caption{Distance estimation errors on BOAT360 for Y7-D (proposed) and Y7-Geo (baseline) under different boat motion conditions: steady (constant speed, straight path) and high dynamic (acceleration, deceleration, and turning).}
\label{tab:poses_distance_errors}
\begin{tabular*}{\columnwidth}{@{\extracolsep{\fill}} c cc cc}
\toprule
\multirow{2}{*}{Movement Condition} & \multicolumn{2}{c}{Y7-Geo} & \multicolumn{2}{c}{Y7-D} \\ 
\cmidrule(lr){2-3} \cmidrule(lr){4-5}
& W.Err & Abs.Err (m) & W.Err & Abs.Err (m) \\ 
\midrule
Steady & 0.609 & 58.629 & 0.299 & \textbf{18.593} \\
High Dynamic & 0.629 & 68.403 & 0.386 & \textbf{32.164} \\
\bottomrule
\end{tabular*}
\vspace{-0.3cm}
\end{table}

Figure \ref{fig:boat_acc} illustratess the impact of extrinsic camera parameter changes on distance estimation for both models. As the boat accelerates, significant pitch angle variations occur. Our proposed learning-based method estimates a distance of 165m, closer to the ground truth of 150m, whereas the geometry method estimates 205m. This demonstrates the robustness of our approach to camera pose changes in dynamic environments. For a more comprehensive comparison, we select 20 images: 10 from scenarios where the boat is traveling in a straight line at a constant speed, and 10 from scenarios where the boat is accelerating, decelerating, and turning. We then analyze the distance prediction errors from both methods, as shown in Table \ref{tab:poses_distance_errors}.

\begin{table*}[htbp]
\centering
\scriptsize
\caption{Distance-wise Weighted and Absolute Errors on the LOAF dataset (baselines italicized).}
\label{tab:LOAF_distance_bins}
\begin{tabular*}{\textwidth}{@{\extracolsep{\fill}} ccc cccc cccc c}
\toprule
\multirow{2}{*}{Distance Bin Range (m)} & \multicolumn{2}{c}{\textit{Y7-Geo}} & \multicolumn{2}{c}{\textit{Y7-DA2}} & \multicolumn{2}{c}{Y7-D} & \multicolumn{2}{c}{Y7-D-Eq} & \multicolumn{2}{c}{Y7-D+C} \\ 
\cmidrule(lr){2-3} \cmidrule(lr){4-5} \cmidrule(lr){6-7} \cmidrule(lr){8-9} \cmidrule(lr){10-11}
& W.Err & Abs.Err (m) & W.Err & Abs.Err (m) & W.Err & Abs.Err (m) & W.Err & Abs.Err (m) & W.Err & Abs.Err (m) \\ 
\midrule
(0.0 - 7.2)  & 1.588 & 8.825  & 0.783 & 4.924 & 0.074 & 0.478 & 0.328 & 1.741 & \textbf{0.058} & \textbf{0.369} \\
(7.2 - 14.4) & 1.109 & 12.203 & 0.894 & 10.132 & \textbf{0.064} &\textbf{ 0.777} & 0.146 & 1.545 & 0.072 & 0.833 \\
(14.4 - 21.6) & 0.671 & 11.081 & 0.928 & 15.829 & \textbf{0.066} &\textbf{ 1.097} & 0.070 & 1.208 & 0.071 & 1.207 \\
(21.6 - 28.8) & 0.510 & 13.350 & 0.952 & 23.372 & 0.095 & 2.179 & 0.090 & \textbf{1.895} & \textbf{0.080 }& 1.974 \\
(28.8 - 36.0) & 0.455 & 14.590 & 0.958 & 27.544 & 0.104 & 3.490 & 0.123 & 3.588 & \textbf{0.075} & \textbf{2.017} \\
\bottomrule
\end{tabular*}
\end{table*}

\begin{table*}[ht]
\centering
\scriptsize
\caption{Object detection and distance estimation performance on ULM360 and BOAT360 (baselines italicized).}
\label{tab:ulm_boat_overall}
\begin{tabular*}{\textwidth}{@{\extracolsep{\fill}} l l cc cccc}
\toprule
Dataset & Model & W.Err & Abs.Err (m) & Precision & Recall & mAP@50 & mAP@50:95 \\
\midrule
\multirow{4}{*}{ULM360} 
& \textit{Y7-w/oDist} & - & - & 0.946 & 0.914 & 0.893 & 0.451 \\
& \textit{Y7-Geo} & 0.396 & 2.443 & 0.963 & 0.897 & 0.908 & 0.525 \\
& \textit{Y7-DA2 }& 0.795 & 5.107 & 0.963 & 0.897 & 0.908 & 0.525  \\
& Y7-D &  \textbf{0.063} &\textbf{0.567} & \textbf{0.963 }& 0.896 & 0.905 & 0.525 \\
& Y7-D-Eq & 0.105& 0.788 & 0.955 & 0.914 & \textbf{0.915} & \textbf{0.539} \\
& Y7-D+C  & 0.078 & 0.645 & 0.955 & 0.819 & 0.855 & 0.438 \\
\midrule
\multirow{4}{*}{BOAT360} 
& \textit{Y7-w/oDist} & - & - & 0.666 & 0.468 & 0.456 & 0.164 \\
& \textit{Y7-Geo} & 0.617 & 65.359 & 0.565 & 0.553 & 0.437 & 0.173 \\
& \textit{Y7-DA2} & 0.980 & 94.255 & 0.565 & 0.553 & 0.437 & 0.173  \\
& Y7-D & \textbf{0.358} & 29.544 & 0.565 & 0.553 & 0.437 & 0.173 \\
& Y7-D-Eq & 0.362 & \textbf{27.935} & \textbf{0.686} & 0.511 & \textbf{0.469} & \textbf{0.175} \\
& Y7-D+C  & 0.447 &29.968 & 0.553 & \textbf{0.54} & 0.423 & 0.154 \\
\bottomrule
\end{tabular*}
\end{table*}

\begin{table*}[ht]
\centering
\scriptsize
\caption{Distance-wise Weighted and Absolute Errors on ULM360 and BOAT360 datasets (baselines italicized).}
\label{tab:ulm_boat_distance_bins}
\begin{tabular*}{\textwidth}{@{\extracolsep{\fill}} c ccc cccc cccc}
\toprule
Dataset & Distance Bin Range (m)  & \multicolumn{2}{c}{\textit{Y7-Geo}} & \multicolumn{2}{c}{\textit{Y7-DA2}} & \multicolumn{2}{c}{Y7-D} & \multicolumn{2}{c}{Y7-D-Eq} & \multicolumn{2}{c}{Y7-D+C} \\
\cline{3-12}
& & W.Err & Abs.Err (m)  & W.Err & Abs.Err (m)  & W.Err & Abs.Err (m) & W.Err & Abs.Err (m) & W.Err & Abs.Err (m) \\
\midrule
\multirow{3}{*}{ULM360} 
& (0.0 - 4.0) & 0.120 & 0.458  & 0.691 & 2.623 & 0.249 & 0.938 & 0.040 & 0.154 & 0.010 & \textbf{0.103} \\
& (4.0 - 8.0) & 0.431 & 2.476  & 0.789 & 4.598 & 0.109 & 0.838 & 0.073 & \textbf{0.657 }& 0.093 & 0.748 \\
& (8.0 - 12.0) & 0.302 & 2.889  & 0.872 & 8.456 & 0.031 & 0.477 & 0.020 & \textbf{0.192} & 0.035 & 0.306 \\
\midrule
\multirow{3}{*}{BOAT360} 
& (0.0 - 100.0) & 0.515 & 28.697  & 0.971 & 54.448 & 0.512 & 23.506 & 0.482 & \textbf{18.487} & 0.667 & 21.424 \\
& (100.0 - 200.0) & 0.723 & 87.459  & 0.989 & 117.872 & 0.188 & \textbf{25.244} & 0.237 & 32.167 & 0.223 & 28.450 \\
& (200.0 - 300.0) & 0.816 & 181.859  & 0.994 & 223.773 & 0.612 & 114.938 & 0.366 & \textbf{91.500} & 0.507 & 99.656 \\
\bottomrule
\end{tabular*}
\vspace{-0.2cm}
\end{table*}

\subsubsection{Learning on Fisheye Images vs.  on Equirectangular Projections}

We compare \textit{Y7-D} (fisheye) and \textit{Y7-D-Eq} (equirectangular)  to assess the impact of the input format. 
As seen in Table~\ref{tab:LOAF_overall}, while \textit{Y7-D-Eq} achieves a higher mAP@50 (0.631 vs. 0.438), its distance estimation error is significantly worse, with an absolute error of 1.609m compared to 0.874m for \textit{Y7-D}. This suggests that training on raw fisheye images provides better depth estimation performance, likely because equirectangular projections distort object shapes and spatial relationships. However, the superior detection accuracy observed in the equirectangular model suggests that object detection benefits from axis-aligned bounding boxes, aligning with the characteristics of datasets such as COCO \cite{cocodataset}, on which YOLO models are pretrained. Importantly, our method, even when trained on equirectangular images, outperforms the geometric and learning-based baselines, demonstrating  its adaptability across projection formats.

\subsubsection{Impact of Camera Height Estimation Head on Distance Prediction}

In traditional geometric methods, such as shown in Figure~\ref{fig:img_lens_model}, camera height is a critical factor for accurate distance estimation. To evaluate whether estimating camera height improves learning-based models, we compare \textit{Y7-D+C} against  \textit{Y7-D}.
Table~\ref{tab:LOAF_overall} shows that estimating camera height (\textit{Y7-D+C}) slightly improves weighted error (0.066 vs. 0.070) but does not  impact absolute distance error. Interestingly, it improves mAP@50 from 0.438 to 0.495, suggesting that incorporating camera height estimation provides additional supervisory signals that enhance object detection. These results indicate that camera height estimation might have acted as a useful auxiliary task, improving both object detection and depth estimation without requiring additional calibration.

To further demonstrate the effectiveness of our approach, we analyze distance estimation performance across different distance ranges. Table~\ref{tab:LOAF_distance_bins} presents the results, illustrating how weighted and absolute errors vary with object distance, further highlighting the superiority of our learning-based method over the baselines.


\subsection{Evaluation on ULM360 and BOAT360 Datasets}

To assess the generalizability of our approach, we evaluate it on the ULM360 and BOAT360 datasets, which feature different sensor setups and environmental conditions, as mentioned in~\ref{subsec:Datasets}.

\subsubsection{Performance on the ULM360 Dataset}

Tables~\ref{tab:ulm_boat_overall} and ~\ref{tab:ulm_boat_distance_bins} present the results on this dataset. Several key observations align with the findings from LOAF:

\begin{itemize}
    \item Learning-based models outperform geometric methods: \textit{Y7-D} achieves an absolute error of 0.57 m, while \textit{Y7-{\text{Geo}}} records a significantly higher error of 2.44 m. 
    
    \item The equirectangular model underperforms in distance estimation: \textit{Y7-D-Eq} shows relatively high detection accuracy (mAP@50 of 0.915) but lags behind \textit{Y7-D} in absolute error (0.567 m vs. 0.788 m).
    
    \item Distance head integration does not degrade object detection: Comparing \textit{Y7-{\text{w/oDist}}}  and \textit{Y7-D}, we observe a slight trade-off in detection recall to obtain distance information.

    \item An additional head estimating the camera height does not enhance the model's performance in detection and distance estimation. A similar trend is observed when the models are evaluated on BOAT360.
\end{itemize}

\subsubsection{Performance on the BOAT360 Dataset}

Tables~\ref{tab:ulm_boat_overall} and~\ref{tab:ulm_boat_distance_bins} report the results on this dataset, which introduces additional challenges such as dynamic motion and greater distance ranges. Key findings include:

\begin{itemize}
    \item Learning-based models outperform geometric methods but show higher errors in extreme conditions: \textit{Y7-D} achieves an absolute error of 29.54 m compared to 65.36 m for \textit{Y7-{\text{Geo}}}. However, compared to LOAF and ULM360, the errors are significantly larger due to the increased dynamics and long-range distance estimation.
    
    \item Distance estimation error increases at longer distances: A notable trend across BOAT360 is that the absolute error grows with increasing distance bins. This suggests that while the learning model remains effective at short to mid-range distances, its performance degrades in long-range estimations. However, compared to our proposed method, the large absolute distance errors of \textit{Y7-{\text{Geo}}} for ranges exceeding 100 meters indicate that the geometric method is unsuitable for mobile platforms with significant dynamics, as variations in camera pose adversely affect its accuracy.

\end{itemize}

\subsection{Summary of Key Findings}

Overall, the experimental results across LOAF, ULM360, and BOAT360 reveal several consistent trends:

\begin{itemize}
    \item Our model maintains object detection accuracy while simultaneously enabling precise distance estimation.
    \item Our proposed learning-based method consistently outperform geometric model and the other learning-based method in distance estimation.
    \item Learning with equirectangular projections improve detection performance but degrade distance estimation.
\end{itemize}

These results underscore the robustness of our learning-based approach, which generalizes well to different environments and sensor setups. While geometric models require precise calibration and are highly sensitive to environmental variations, our deep learning model adapts effectively to diverse conditions.



\section{CONCLUSION}
In this work, we introduced a neural learning-based approach for real-time monocular distance estimation using omnidirectional fisheye images. Our method eliminates the need for precise lens calibration and explicitly models depth relationships, making it more robust to real-world variations such as lens distortions, dynamic movements, and environmental changes.

Through extensive experiments on three challenging datasets (LOAF, ULM360, and BOAT360) we demonstrated that our learning-based model consistently outperforms traditional geometric methods in both accuracy and robustness. In particular, our approach showed improved resilience to camera pose variations and extrinsic parameter shifts, a critical factor for deployment in dynamic environments such as moving vehicles and boats.

Future work will explore hybrid models that integrate geometric priors into deep learning architectures, improving generalization to diverse real-world applications. Furthermore, expanding our training datasets with more varied dynamic scenes will enhance model robustness for practical deployment in robotics, autonomous navigation, and surveillance.

\bibliographystyle{./IEEEtran} 
\bibliography{./IEEEabrv,./IEEEexample}

\end{document}